%% file: ijcai24.tex
\documentclass{article}
\usepackage{ijcai24}
\usepackage{times}
\usepackage{soul}
\usepackage{url}
\usepackage{xcolor}
\usepackage[hidelinks]{hyperref}
\usepackage[utf8]{inputenc}
\usepackage[small]{caption}
\usepackage{graphicx,subfloat}
\usepackage{amsthm, amsmath, amssymb}
\usepackage{booktabs}
\usepackage{algorithm}
\usepackage{algorithmic}
\usepackage[switch]{lineno}
\usepackage{arydshln}
\usepackage{bm}
\usepackage{color}
\usepackage{subcaption}
\usepackage{microtype}

\title{Dynamically Anchored Prompting for Task-Imbalanced Continual Learning}

\author{
Chenxing Hong$^{1}$
\and
Yan Jin$^{1,2}$\and
Zhiqi Kang$^4$\and
Yizhou Chen$^{1}$\and
Mengke Li$^3$\and
Yang Lu$^{1,2}$\footnote{Corresponding Author: Yang Lu (luyang@xmu.edu.cn)}\And
\\Hanzi Wang$^{1,2}$\\
\affiliations
$^1$Key Laboratory of Multimedia Trusted Perception and Efficient Computing, Ministry of Education of China, Xiamen University, Xiamen, China\\
$^2$Fujian Key Laboratory of Sensing and Computing for Smart City, School of Informatics, Xiamen University, Xiamen, China\\
$^3$Guangdong Laboratory of Artificial Intelligence and Digital Economy (SZ), Shenzhen, China\\
$^4$Univ. Grenoble Alpes, Inria, CNRS, Grenoble INP, LJK, 38000 Grenoble, France.\\
\emails
hongchenxing@stu.xmu.edu.cn,
jinyan7973@gmail.com,
zhiqi.kang@inria.fr,
36920231153184@stu.xmu.edu.cn,
limengke@gml.ac.cn,
\{luyang, hanzi.wang\}@xmu.edu.cn
}

\begin{document}

\maketitle

\begin{abstract}
Existing continual learning literature relies heavily on a strong assumption that tasks arrive with a balanced data stream, which is often unrealistic in real-world applications. 
In this work, we explore task-imbalanced continual learning (TICL) scenarios where 
the distribution of task data is non-uniform across the whole learning process.
We find that imbalanced tasks significantly challenge the capability of models to control the trade-off between stability and plasticity from the perspective of recent prompt-based continual learning methods. 
On top of the above finding, we propose Dynamically Anchored Prompting (DAP), a prompt-based method that only maintains a single general prompt to adapt to the shifts within a task stream dynamically.
This general prompt is regularized in the prompt space with two specifically designed prompt anchors, called boosting anchor and stabilizing anchor, to balance stability and plasticity in TICL.
Remarkably, DAP achieves this balance by only storing a prompt across the data stream, therefore offering a substantial advantage in rehearsal-free CL.
Extensive experiments demonstrate that the proposed DAP results in 4.5\% to 15\% absolute improvements over state-of-the-art methods on benchmarks under task-imbalanced settings. 
Our code is available at {\href{https://github.com/chenxing6666/DAP}{{https://github.com/chenxing6666/DAP}}}.

\end{abstract}

\section{Introduction}


Human beings possess the remarkable ability to learn new tasks and solve evolving challenges by leveraging knowledge from their past experiences. 
Inspired by this, continual learning (CL) methods are designed to address a series of tasks using a singular model while preserving performance on tasks previously mastered~\cite{wang2022continual,zhang2020class,aljundi2018memory}.
However, achieving this goal is challenging for deep models, as they tend to easily forget previously learned information, i.e., a phenomenon known as catastrophic forgetting~\cite{de2021continual,kirkpatrick2017overcoming}. 
This issue primarily arises from the network's tendency to overwrite old knowledge with new data during the training process.

\input{figures/introduction}

Although existing methods in CL have achieved notable progress, they generally assume a balanced distribution of training data across tasks, i.e., each task holds the same number of training samples.
In practical applications, data streams often exhibit an imbalanced distribution~\cite{huang2019deep}, where the data volume for each task can vary significantly, with some tasks presenting a large number of samples while others may have far fewer. 
This gives rise to task-imbalanced CL scenarios and introduces potential new issues for the CL process. 
The imbalance among tasks will amplify the difficulty of striking a balance between learning and retaining knowledge over time. 

To address the aforementioned problem, this paper first investigates this more general and realistic scenario: task-imbalanced continual learning (TICL). 
TICL characterizes environments where the number of samples in each class is imbalanced across the data stream, reflecting the long-tailed nature observed in most real-world data distributions.
In such distributions, a few classes dominate in terms of the number of samples, while many others have significantly fewer. 
In the context of CL, this imbalance manifests at the task level, leading to imbalanced tasks, as shown in  Figure~\ref{fig:intro}(a). 
We call the tasks with a relatively large number of samples the \textit{data-rich tasks} and the tasks with a relatively small number of samples the \textit{data-scarce tasks}.
Specifically, we consider three cases of TICL: 
\begin{itemize}
\item \textbf{Descending TICL}: The learning process starts with data-rich tasks followed by data-scarce ones.
\item \textbf{Ascending TICL}: The learning process starts with data-scarce tasks preceding data-rich ones.
\item \textbf{Shuffled TICL}: Tasks arrive in a random sequence without a prescribed order.
\end{itemize}
While descending and ascending TICL are two extreme cases, shuffled TICL can be regarded as the general form of TICL.

To show the challenges brought by the proposed scenarios, we conduct a quick experiment using DualPrompt~\cite{wang2022dualprompt}, a typical prompt-based CL method, in Figure~\ref{fig:intro}(b). It can be observed that in the descending TICL case, the model initially learns well on data-rich tasks but rapidly declines when training becomes scarce.
This indicates the poor plasticity of the model in adapting to new tasks with fewer samples. 
On the contrary, in the ascending TICL case, the model initially struggles with learning data-scarce tasks and remains poor in the following tasks.
This is due to the severe forgetting caused by the following data-rich tasks (see Figure.~\ref{fig:motivation} for more analysis).
These results reveal a more realistic plasticity and stability dilemma: given an incoming task, the model learns poorly when its training data is scarce and forgets rapidly when there is abundant data. 
Undoubtedly, both learning and forgetting become more challenging in TICL. 
Therefore, there is a clear need to design a specific method to address this issue.

In this paper, we propose a novel prompt-based approach named Dynamically Anchored Prompting (DAP).
DAP only maintains a general prompt to learn from imbalanced task streams by strategically address the dilemma of stability and plasticity.
It tackles the dilemma by decoupling stability and plasticity through two specialized prompts: the boosting prompt and the stabilizing prompt.
The stabilizing prompt focuses on preserving the knowledge of past tasks, thereby ensuring stability and mitigating catastrophic forgetting. 
In contrast, the boosting prompt enhances the model's generalization ability to learn and adapt to new tasks, promoting plasticity. 
The general prompt is updated by a novel dynamic stability-plasticity regularization (DSPR) strategy, which dynamically regularizes the general prompt in the prompt space based on task attributes, ensuring a flexible and adaptive learning process.
Since DAP only stores a general prompt, it achieves superior performance with lower memory requirements, aligning well with the objectives of rehearsal-free CL. 
In summary, our contributions are three-fold:

\begin{itemize}

    \item We analyze a more realistic CL scenario with imbalanced tasks along with prompt-based learning algorithms, uncovering their defects caused by the stability and plasticity dilemma.
    This dilemma highlights a critical challenge: an effective balance between preserving existing knowledge and accommodating new learning demands.
    \item We propose DAP, a novel approach to dynamically balance the stability and plasticity with a single regularized general prompt, which effectively addresses the challenges in TICL. 
    \item We evaluate the performance of DAP on benchmark datasets. Our proposed DAP exhibits significant improvements over previous state-of-the-art methods, with large margins ranging from 4.5\% to 15\%.
\end{itemize}

\section{Related Work}

\subsection{Continual Learning}

The existing research in CL within machine learning primarily assumes a balanced task distribution and focuses on addressing catastrophic forgetting, which can be categorized into three strategies~\cite{de2021continual,wang2023comprehensive}: architectural expansion, regularization, and rehearsal. 
Architectural expansion adapts the model's structure for new tasks, suitable where adding to the model is practical~\cite{kang2022forget,wang2022coscl}. 
Regularization, either in the weight or prediction space, aims to retain past task knowledge during new task training, with knowledge distillation being particularly effective in prediction space~\cite{castro2018end,cha2021co2l}. 
Rehearsal methods, using original or synthetic data, are efficient but face data privacy and storage issues~\cite{chaudhry2021using,wang2022foster,kang2023soft}. 
These issues emphasize the need for rehearsal-free methods in CL, addressing both privacy concerns and computational efficiency~\cite{vaishnavh2018theoretical}.

\subsection{Prompting for Continual Learning}
The recent trend in CL research focuses on combining prompting techniques with Vision Transformers (ViTs). 
This approach involves using a pre-trained, frozen backbone model from ImageNet, circumventing the need for a replay buffer. 
Prompting, initially applied in transfer learning with pre-trained language models like GPT-3, involves adding language-based instructions to the input text to guide the model in understanding downstream tasks. 
Traditional prompting methods were heuristic, but recent developments like Prompt Tuning~\cite{kemker2017fearnet} and Prefix Tuning~\cite{kemker2018measuring} introduced the concept of learnable prompts in a continuous space, becoming mainstream in prompt-based learning.

In the realm of prompt-based CL, various methods have been proposed. Specifically,
L2P~\cite{wang2022learning} introduced a prompt pool concept to adjust the frozen ViT backbone for CL tasks. Building on this, DualPrompt~\cite{wang2022dualprompt} employs two different prompt types: G-Prompt for learning task-invariant knowledge and E-Prompt for task-specific knowledge, drawing inspiration from complementary learning systems.
CODA-Prompt~\cite{smith2023coda} adopts input-conditioned prompts through an innovative attention-based end-to-end key-query mechanism that integrates the entire training sequence.

\subsection{Imbalanced Continual Learning}
While there is existing research on addressing imbalance in CL, it primarily concentrates on specific aspects of imbalance. For instance, BIC~\cite{wu2019large} focuses on the imbalance between limited stored samples and current task samples, addressing challenges in storage under highly imbalanced conditions. PRS~\cite{kim2020imbalanced} investigates the long-tail problem in multi-label scenarios, also relying on sample storage. Two-Stage-CL~\cite{liu2022long} delves into long-tail class imbalances in CL, storing substantial amounts of original data.

It is noteworthy that these studies are primarily rehearsal-based methods. 
This underscores the significant challenge of storing samples where class imbalance is inherent, as data selection itself is imbalanced.
Furthermore, our approach to TICL diverges fundamentally from these existing works. 
While they primarily concentrate on addressing imbalances within samples or classes, our focus lies in the imbalance across different tasks.

\section{Problem Formulation}
\subsection{Preliminaries}

Our CL protocol adopts the class-incremental CL setting.
The training data is denoted as a sequence of $T$ tasks $\mathcal{D} = \{\mathcal{D}_0, \ldots, \mathcal{D}_T\}$, where $\mathcal{D}_t = \{(x_i^t, y_i^t)\}, i=0,...,N_t$ is sampled from a joint data distribution in the input and label space $X_t \times Y_t$ at task $t$, whose size (i.e., the number of samples in this task) is denoted as $N_t$.
The target model is formulated as $f : X \rightarrow Y$, integrating a patch embedding layer $f_p$, and a backbone $f_b$ consisting of a stack of transformer encoder layers followed by a classifier, thus $f = f_p \circ f_b$. 
We employ a ViT-Base model, pre-trained on ImageNet, as the frozen feature extractor.
In this class-incremental setting, similar to recent prompt-based CL methods\cite{smith2023coda,wang2022dualprompt,wang2022learning}, task boundaries are clearly defined with no shared classes between them, and task identity is provided only during training.

\subsection{Task-Imbalanced Continual Learning}

In this paper, we formally study task-imbalanced continual learning (TICL).
TICL is characterized by the distribution of task sizes, i.e., $N_t$, throughout the learning process. 
Different from the conventional task-balanced CL where $N_t$ is drawn from a uniform distribution, i.e., $N_i=N_j, \forall i\ne j~\text{and}~i,j\in \{0, ..., T\}$. 
Any distribution deviating from the uniform one introduces an inherent imbalance among tasks.
By the long-tail nature of real-world data distributions, we generally assume that the task sizes also follow the long-tail distribution. 
Specifically, the task sizes follow $N_{I_0} > ... > N_{I_T},~\forall I_i<I_j$, where $I_0,...,I_T$ are the sorting index based on the task size by non-increasing order.

\subsection{Case Study}

\input{figures/motivation_fig.tex}

We first study how deep the model performance might drop on two extreme cases of TICL: descending TICL and ascending TICL.
To properly quantify plasticity and stability~{\cite{sun2022exploring}}, we adopt two metrics to evaluate a model on the task $t$:
\begin{equation}
P = Acc_{t,t}, \quad F =  Acc_{t,t} - Acc_{T,t}.
\end{equation}
where $Acc_{i,j}$ represents the accuracy on the test set of the task $j$ after finishing training on the task $i$.
$P$ measures the models' plasticity, i.e., the ability to learn new tasks, while $F$ measures the model's stability, i.e., the resistance to catastrophic forgetting. 
Here, we examine the performance of DualPrompt~\cite{wang2022dualprompt} on descending TICL and ascending TICL with evaluation metrics $P$ and $F$.
As a recent prompt-based CL method, it achieves a good balance between stability and plasticity in task-balanced scenarios.
In addition, to eliminate the factor of the absolutely small size of each task and focus on the task imbalance problem, we also compare with the case of one-shot, where each task contains only one sample for each class.

\input{figures/method_fig.tex}

\noindent\textbf{Stability.} 
Stability is extremely hard to achieve by a model trained on ascending TICL as catastrophic forgetting occurs more easily on past data-scarce tasks.
As shown in Figure \ref{fig:motivation}(a), 
The model's performance decline on earlier tasks in the ascending case is significant, which indicates severe catastrophic forgetting. 
The result of the one-shot case reveals that this forgetting is not just due to limited data quantity. 
On the contrary, in the descending case, the model keeps good stability because the data-rich tasks come first, and the following data-scarce tasks can hardly overtake the knowledge of the data-rich tasks in the model.

\noindent\textbf{Plasticity.} 
On the opposite, plasticity is also extremely hard to achieve by a model trained on descending TICL as the model learns well on past data-rich tasks such that the following data-scarce tasks cannot be generalized well.
As shown in Figure \ref{fig:motivation}(b), the model performance in the descending case significantly drops.
This raises the issue that the model initially trained on data-rich tasks struggles to adapt to the following data-scarce tasks, prioritizing stability over learning new information.
We also compare with the one-shot case.
The model shows consistent learning performance starting from the fifth task, surprisingly suggesting better adaptation if the last few tasks are not data-rich.

Accordingly, we propose a method to effectively balance the stability-plasticity dilemma in TICL scenarios.

\section{Dynamically Anchored Prompting}
To address the issues identified in the case study, we propose Dynamically Anchored Prompting (DAP) for TICL in this paper. 
The proposed method is designed to dynamically balance and optimize the trade-off between stability and plasticity, empowering the model flexible to learn in TICL.

Different from the existing prompt-based CL methods~\cite{wang2022dualprompt,smith2023coda,wang2022learning}, the key idea of DAP is to maintain a prompt across all the tasks, which is called the \textit{general prompt}.
The general prompt aims to learn to generalize each task with the knowledge of the pre-trained model.
It can be easily used during inference as there is no prompt pool, and therefore no prompt selection is needed.
During the inference, the challenge of adopting only a general prompt is to balance the knowledge learned from each task, especially for imbalanced tasks.
Therefore, to achieve the goal of using the general prompt to generalize well across all tasks, the proposed DAP adopts a two-phase learning scheme for each training task, termed in-task phases.
Given a specific training task, during the first in-task phase, we optimize a \textit{task-specific prompt}, which is then utilized to regularize the general prompt during the second in-task phase.
To tackle the imbalance problem of TICL, we propose dynamic stability-plasticity regularization to make the general prompt learn well and stably no matter the amount of training data in the next incoming task.
After two-phase tuning for each task, we only need to save the general prompt for inference.
The overall framework of the proposed DAP is shown in Figure \ref{fig:method}.

\subsection{Anchored Prompting}
The goal of DAP is to obtain a high-quality general prompt that properly addresses the intrinsic problem of TICL. 
For each task, we first optimize the task-specific prompt in the first in-task phase.
The task-specific prompt is used to form two key components: the \textit{boosting anchor} $\mathbf{p}_b$ and the \textit{stabilizing anchor} $\mathbf{p}_s$. 
These anchors are designed to regularize the relationship between the \textit{general prompt} $\mathbf{p}_g$ and each anchor, effectively handling the plasticity and stability dilemma during training. 
The general prompt is then optimized in the prompt space with a new strategy called dynamic stability-plasticity regularization.

\noindent\textbf{Boosting Anchor.}  
Specifically, the boosting anchor $\mathbf{p}_b$ aims to maintain model plasticity, ensuring adaptability to new tasks. 
It is especially useful when the size of the current task is small compared to the past tasks. 
The boosting anchor $\mathbf{p}_b$ is simply set at the task-specific prompt optimized on the current task to capture the task-relevant information. 
Thus, $\mathbf{p}_b$ functions as a critical focal point, guiding the model towards learning trajectories that maximize plasticity.
To optimize $\mathbf{p}_b$, we formulate the following loss function:
\begin{equation}
\mathcal{L}_1 = \sum_{i=0}^{N_t}l_{CE}(f_b([\mathbf{p}_b, f_p(x_{i}^{t})]), y_{i}^{t})
\end{equation}
Here, $h(x_{i}^{t})$ and $y_{i}^{t}$ represent the patched features of the $i$-th sample and its corresponding label from task $t$, respectively.
$\mathbf{p}_b$ is initialized in each new task.
The concatenation $[\mathbf{p}_b, h(x_{i}^{t})]$ is fed into the pre-trained transformer as the input.
$\mathcal{L}_1$ represents the total loss for the first in-task phase, and $l_{CE}$ is the cross-entropy loss.
The goal of minimizing this loss is to make $\mathbf{p}_b$ fully learn the knowledge for adapting the pre-trained model for the current task.

\noindent\textbf{Stabilizing Anchor.} 
To ensure model stability, the stabilizing anchor $\mathbf{p}_s$ is designed to prevent knowledge forgetting from past tasks by monitoring the learned task-specific prompts.
To maintain the knowledge of all of the learned tasks, $\mathbf{p}_s$ is calculated by the weighted center of the boosting anchors of all learned tasks in the prompt space from task 0 to the current task $t$.
The weight is associated with the inverse of the size of each task.
To make the algorithm rehearsal-free, i.e., without storing any data or prompt, $\mathbf{p}_s$ can be updated in an online manner:
\begin{align}\label{stabilizing_anchor}
\mathbf{p}_s \leftarrow \frac{1}{\sum_{i=0}^{t}\frac{1}{N_i}}\left({\sum_{i=0}^{t-1}\frac{\mathbf{p}_s}{N_i}} + \frac{\mathbf{p}_b}{N_{t}}\right).
\end{align}
The rationale behind this update is that task-specific prompts are optimized to generalize well in the corresponding task, thus, their weighted average represents stability.
The weights in Eq.~(\ref{stabilizing_anchor}) associated with the corresponding task size $N_i$ emphasize the past task with a smaller size to make the stabilizing prompt uniformly represent the knowledge of past tasks.

\noindent\textbf{Anchor Alignment.} 
The purpose of maintaining the boosting anchor and the stabilizing anchor in each task is to dynamically regularize the learning of the general prompt, such that it can be flexible to balance the stability and plasticity in the TICL scenario.
We employ cosine similarity to measure the proximity between the general prompt and anchors in each task for anchor alignment, as it focuses on the orientation rather than the magnitude of vector representations, effectively capturing the inherent relationships between prompts:
\begin{equation}
    \mathcal{L}_a(\mathbf{p}_g, \mathbf{p}) = 1 - \frac{\mathbf{p}_g \cdot \mathbf{p}}{\|\mathbf{p}_g\|\|\mathbf{p}\|},
\end{equation}
where $\mathbf{p}$ can be either $\mathbf{p}_b$ or $\mathbf{p}_s$ to address the plasticity or the stability, respectively.
Different from the methods that simultaneously adopt two sets of prompts~\cite{wang2022dualprompt}, the proposed DAP only utilizes the boosting anchors (i.e. the task-specific prompts) as a constant to regularize the general prompt by anchor alignment.
Therefore, task-specific prompts are not involved in the final inference process.
The advantages of only maintaining the general prompt thought the CL process are two-fold. 
On one hand, it avoids the error produced by matching the improper prompts in the prompt pool~\cite{wang2022learning}.
On the other hand, it makes the proposed DAP fully rehearsal-free.

\subsection{Dynamic Stability-Plasticity Regularization}
Anchored prompting offers an opportunity to address either plasticity or stability individually, yet it falls short in dynamically adapting within complex, imbalanced task streams. 
Recall the three cases in TICL described in Figure~\ref{fig:intro}(a).
Updating the general prompt by anchor alignment with a single anchor may only work for the descending TICL or the ascending TICL.
For shuffled TICL, the incoming task cannot be guaranteed to be a data-scarce or data-rich task, such that the emphasis on updating the general prompt cannot be determined.
The challenge of achieving balance in such fluctuating scenarios remains unresolved.

To address the remaining issue in the DAP framework, we introduce a strategy that enables the model to adjust its focus between plasticity and stability in the nature of TICL tasks.
In the second in-task phase, we optimize the general prompt $\mathbf{p}_g$ by considering the balance between stability and plasticity.
Different from $\mathbf{p}_b$ that is initialized in each new task, $\mathbf{p}_g$ is initialized in task 0 and updated through the whole CL process in order to achieve the ability to generalize well on all learned classes.
Thus, we propose dynamic stability-plasticity factor $\lambda$ as the coefficient between two anchor alignments. 
The factor $\lambda$ modulates the balance between stability and plasticity by considering the size of the current task $t$ and the sizes of past tasks:
\begin{equation}
\lambda = \frac{N_t - N_{\min}}{N_{\max} - N_{\min} + \epsilon},
\end{equation}
where $N_{\min}=\min\{N_i\} $ and $N_{\max} = \max \{N_i\}$ are the minimum and maximum sizes of learned tasks for $i=0,...,t$, and $\epsilon$ is a small positive constant to prevent division by zero. 
It is basically the min-max normalization that measures the difference between the size of the current task and the minimum size of the learned tasks, which is then normalized into the range of $[0,1]$.
As with many re-weighting techniques in long-tail learning~\cite{peng2023dynamic,chen2023area,zhang2023deep}, the task size $N_t$ is an important indicator to represent the learning difficulty.
It is relatively easier to learn from a task with a larger size.
Accordingly, we use the factor $\lambda$ and $1-\lambda$ as the regularization coefficient to reflect this relationship.
In the second in-task phase, the general prompt updated in the current task $t$ is given by the following loss function with the dynamic stability-plasticity factor $\lambda$:
\begin{align}\nonumber
\mathcal{L}_2 = &\sum_{i=0}^{N_t}l_{CE}(f_b([\mathbf{p}_g, f_p(x_{i}^{t})]), y_{i}^{t}) \\
&+ \lambda \cdot \mathcal{L}_a(\mathbf{p}_g, \mathbf{p}_s) + ( 1 - \lambda ) \cdot \mathcal{L}_a(\mathbf{p}_g, \mathbf{p}_b).
\end{align}
Therefore, a larger $N_t$ indicates a smaller $\lambda$, enhancing stability to prevent forgetting. 
On the other hand, a smaller $N_n$, indicative of a more challenging task, prompts an increase in $\lambda$ to ensure sufficient plasticity for learning new, complex tasks. 
If the size of the current task $t$ is the largest or smallest ever, $\lambda$ then becomes 1 or 0, respectively.

The effective utilization of the prompt anchors and the flexible adjustment of $\lambda$ according to the task sizes are the core of DAP, which adapts the pre-trained model well in the scenario of TICL.
It facilitates a balance between the acquisition of new knowledge and the retention of prior learning.

\input{tables/table.tex}

\section{Experimental Results}
In this section, we first introduce the experimental setup and then compare the proposed DAP with different existing CL methods applied on TICL benchmarks. 
Finally, we specifically evaluate the effectiveness of DAP and conduct ablation studies to key elements.

\subsection{Implementation Details}
\noindent\textbf{Datasets.}
Given that long-tail distributions are the most prevalent form of imbalance in the real-world, we adopt the long-tailed setting to construct imbalances.
The long-tailed distribution typically follows an exponential decay in sample size across classes~\cite{cao2019learning}. 
This decay is parameterized by $\rho$ which is the ratio between the most and least frequent classes.  $\rho=1$ is the conventional CIL case and $\rho$ in (0,1) indicates different degrees of long-tailed distribution.

We follow the experimental datasets used in previous works~\cite{wang2022dualprompt,khan2023introducing}, first conducting a long-tail division of the datasets, and then dividing them into 10 disjoint tasks. Specifically, the CIFAR-100 dataset~\cite{krizhevsky2009learning} includes 100 classes of natural images, with 500 training samples for the head classes, and subsequently decreasing for the remaining classes according to the long-tail division method. 
To ensure balance within each task, we select an equal number of samples from each class within a task, based on the maximum class quantity present in that task. 
The ImageNet-R dataset~\cite{wang2022dualprompt} contains 200 classes of images, divided in a similar manner for long-tail calculation.
Aligned with this long-tailed distribution approach, we examine three cases in our study: Descending TICL, where learners first encounter data-rich tasks followed by data-scarce ones; Ascending TICL, featuring data-scarce tasks preceding data-rich ones; Shuffled TICL, where tasks arrive in a random sequence without a prescribed order of data volume.

\noindent\textbf{Comparison Methods.} 
In our experiments, we compare the proposed DAP with two groups of rehearsal-based methods. The first group includes classical approaches like PODNET~\cite{douillard2020podnet} and BiC~\cite{wu2019large}. The second group comprises methods designed for long-tailed distributions such as EEIF++~\cite{liu2022long} and LUCIR++~\cite{liu2022long}.
Additionally, we include pretrained methods in our comparison, FineTune~\cite{khan2023introducing}, iCaRL~\cite{rebuffi2017icarl}, both these methods start from the same ImageNet pre-trained ViT-Base~\cite{dosovitskiy2020image} model to ensure a fair comparison.
Lastly, we compare DAP against the current state-of-the-art (SOTA) prompt-based methods, including L2P~\cite{wang2022learning}, DualPrompt~\cite{wang2022dualprompt}, and CODA-Prompt~\cite{smith2023coda}.

\noindent\textbf{Evaluation Protocol.} 
For the test set, we followed the setting of long-tailed learning research \cite{cao2019learning} where the training set is imbalanced while the test set is balanced. 
Therefore, our test set is consistent with the balanced CL~\cite{wang2022continual,wang2022dualprompt}. 
In this manner, the testing accuracy can be easily averaged over all classes to reflect the performance of each class with equal weight.
For evaluation, we report their average values with standard errors using two widely used CL metrics: average accuracy (\textit{A\textsubscript{N}} $\uparrow$) \cite{lopez2017discovering} of the final average accuracy by the model, last accuracy (\textit{A\textsubscript{L}} $\uparrow$) \cite{zhang2023slca} of the last accuracy at the end of the learning process.

\noindent\textbf{Implementation.}
Following the settings of L2P~\cite{wang2022learning}, We train DAP using Adam with $\beta_1, \beta_2$ of 0.9, a learning rate of 0.01, and a batch size of 64. We resize the input images to a 224$\times$224 resolution and normalize them between 0 and 1. To ensure models converge, we train TICL-CIFAR-100 for 5 epochs per task, TICL-ImageNet-R for 50 epochs each task.

\subsection{Comparison to the State-of-The-Art}

We compare various rehearsal-based and prompt-based methods for TICL-CIFAR-100 and TICL-ImageNet-R in Table~\ref{tab:cifar100_main}. 
We observe that DAP consistently outperforms all rehearsal-based methods by a considerable margin, with a substantial improvement ranging from 10\% to 30\%, establishing a new state-of-the-art in all cases.
It also demonstrates a significant advantage over other prompt-based methods. showing an increase of 4.5\% to 15\%.

\subsection{Effectiveness of the General Prompt}
To verify that adopting a single general prompt $\mathbf{p}_g$ is able to accumulate knowledge across the data stream, we adopt a linear probing experiment ~\cite{he2020momentum} to evaluate the performance of the representation layer. 
Specifically, following each incremental learning task, we freeze the representation layer and introduce an additional classification layer known as a linear probe, which is trained on all classes of the benchmark dataset.

We conducted a detailed analysis of the descending, ascending, and shuffled cases. 
As illustrated in Figure~\ref{fig:present}, in the descending case, we observe a rapid initial improvement, indicating that the model indeed learns significant global knowledge when presented with abundant data initially. 
However, as the tasks progress and data availability decreases, we notice a stagnation in learning, suggesting that under typical conditions, the model struggles to acquire new knowledge with limited data. 
In contrast, DAP continues to improve even with less data, overcoming this learning stagnation. 
In the ascending case, despite continuous learning, the final performance is lower than in the descending case, implying some degree of knowledge forgetting. 
Yet, DAP still shows an upward trend at the end. 
A similar pattern is observed in the shuffled case.
This demonstrates DAP successfully accumulated knowledge, effectively navigating the numerical disparities in TICL environments.
Moreover, DAP consistently outperforms DualPrompt~\cite{wang2022dualprompt} through all the tasks.

\input{figures/model_eval_pre}

\subsection{Ablation Study}

In this section, we delve into an in-depth ablation study to validate the effectiveness and contributions of different components in our model.

\input{tables/abla_g_and_t}

\noindent\textbf{Dynamic Factor $\lambda$.}
 $\lambda$ is essential for calibrating the balance between stability and plasticity.
Therefore, to assess the dynamic regularization's effectiveness, we compare the results in the shuffled case to the use of the fixed values of $\lambda$ from 0 to 1.
As illustrated in Figure~\ref{fig:aba_lamda}, the dynamically adjusted $\lambda$ consistently outperforms any fixed value of $\lambda$.
This supports our premise that a dynamic $\lambda$ is more flexible to adapt to the evolving requirements of learning new information while retaining previously acquired knowledge.
Generalization on all classes is ensured irrespective of the data abundance in each task during learning.

\noindent\textbf{Task-specific Prompt and General Prompt.}
We further examine the roles of prompts within the DAP framework. 
Since DAP optimizes the task-specific prompt for each task in the first in-task phase and continually updates the general prompt in the second in-task phase, we can investigate the performance of using either prompt exclusively. 
This purpose of the study aims to discern whether each prompt can sustain the model's efficacy across various learning scenarios on its own.
We compare DAP with two exclusively designed methods: 
(1) To solely use the task-specific prompt. 
The task-specific prompts for all tasks are stored during inference, and it is assumed that the prompt is properly selected for each test sample.
(2) To solely use the general prompt. The general prompt is continually optimized without dynamic stability-plasticity regularization.
As shown in Table~\ref{tab:abla_g_t}, employing either the task-specific prompt or the general prompt solely yields results that are much inferior to DAP. 
The task-specific prompt cannot harness the information across tasks while updating the general prompt without regularization cannot well balance the stability and plasticity. 
The strength of DAP's design of updating the general prompt with regularization is verified in addressing the demands of TICL.

\input{figures/abla_lamda}

\input{tables/abla_anchor}

\noindent\textbf{Ablation on Anchors.} To ablate the boosting anchor and stabilizing anchor, we focused exclusively on the shuffled case of TICL, because this case uniquely demands the model to effectively balance both stability and plasticity.
Therefore, we conducted experiments in this setting using each anchor type in isolation without the dynamic stability-plasticity regularization, aiming to assess the capability of each anchor.
As seen from Table~\ref{tab:abla_anchor}, employing the boosting anchor or the stabilizing anchor in isolation only enhances the model's plasticity or stability but at a significant cost to each other. 
The best performance is attained by combining both boosting and stabilizing anchors with the dynamic factor.

\section{Conclusion}
In this paper, we formally define task-imbalanced continual learning and systematically study its three cases. 
We discovered that imbalanced tasks significantly deteriorate the performance of prompt-based CL methods because they raise a new challenge to consider the dilemma of stability and plasticity. 
To counteract this, we introduced dynamically anchored prompting.
DAP addresses the challenge by separating stability and plasticity with two prompts, one for stabilization and the other for plasticity, serving as anchors to guide the learning process of a general prompt. 
DAP improves the performance of baseline prompt-based TICL methods to set a new state-of-the-art.

\section*{Acknowledgements}
\textls[-5]{This study was supported in part by the National Natural Science Foundation of China under Grants 62376233,\ U21A20514,\ and\ 62306181;\ in\ part\ by\ the FuXiaQuan National Independent Innovation Demonstration Zone\ Collaborative\ Innovation\ Platform\ under\ Grant\ 3502ZCQXT2022008; in part by the Natural Science Foundation of Guangdong Province under Grant 2024A1515010163; in part by the China Fundamental Research Funds for the Central Universities under Grant 20720230038; and in part by Xiaomi Young Talents Program.}



\end{document}

%% file: figures/introduction.tex
\begin{figure}[!t]
\subfloat[TICL cases]{ 
    \includegraphics[width=0.47\linewidth]{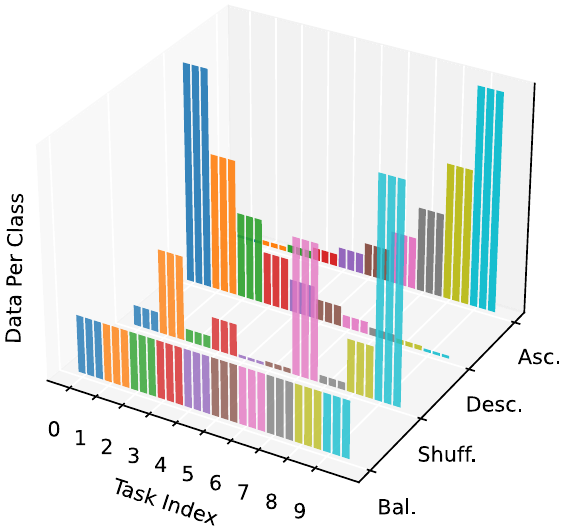}
    \label{fig:TICL_dis}
    }
\subfloat[Performance degradation ]{
    \includegraphics[width=0.46\linewidth]{figures/avg_accuracy_lineplot-2b.pdf}
    \label{fig:perf_deg}
    }    
\caption{An illustration of the scenario of task-imbalanced continual learning (TICL).  (a) Three cases in TICL compared to ordinary task-balanced CL. (b) Performance degradation of DualPrompt on TICL CIFAR-100. The number of balanced and imbalanced tasks is ensured to be the same.}
\label{fig:intro}
\vspace{-5pt}
\end{figure}

%% file: figures/motivation_fig.tex
\begin{figure}[t]
  \centering
    \begin{minipage}{0.5\linewidth}
    \centering
    \includegraphics[width=0.95\linewidth]{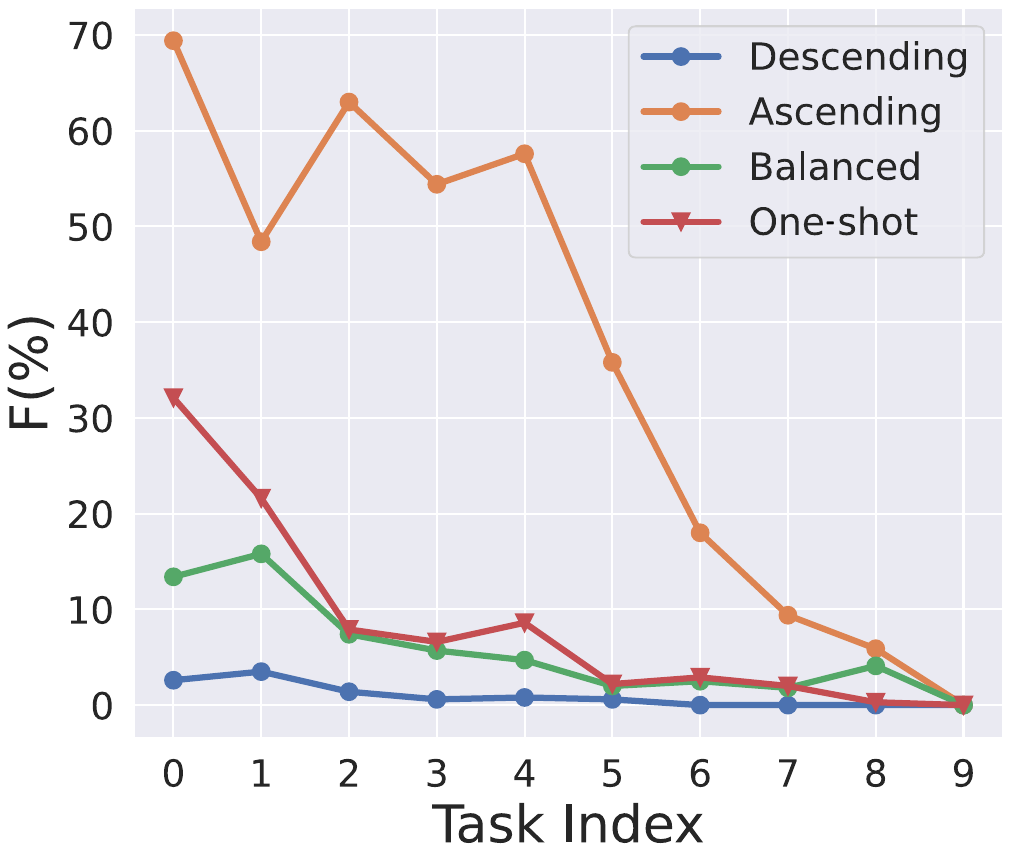}
   \subcaption{Stability}
    \label{fig:sta_perf}
  \end{minipage}\hfill
  \begin{minipage}{0.5\linewidth}
    \centering
    \includegraphics[width=0.95\linewidth]{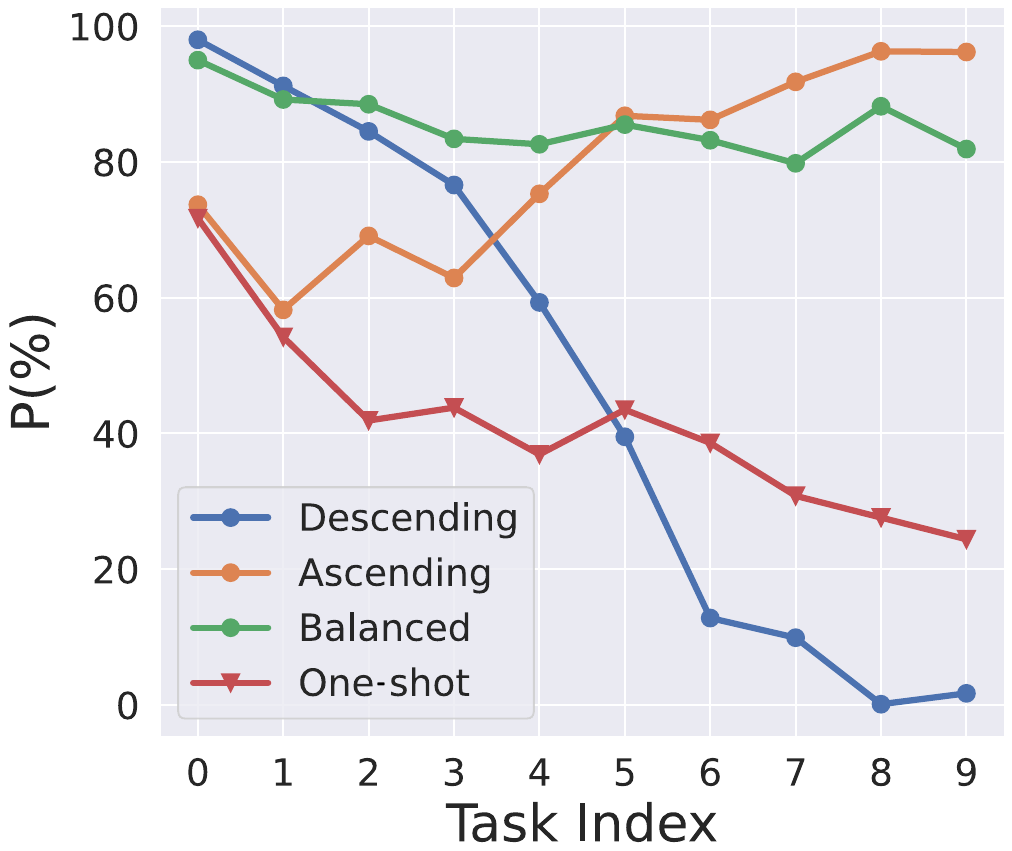}
    \subcaption{Plasticity }
    \label{fig:pla_perf}
  \end{minipage}
  
  \caption{Performance analysis of DualPrompt on different TICL cases with evaluation metrics. \small ({a}) $F$ shows the model stability to resist forgetting. ({b}) $P$ shows the model plasticity to learn new tasks.}\label{fig:motivation}
  \vspace{-3.5mm}
\end{figure}

%% file: figures/method_fig.tex
\begin{figure*}[t]
    \centering
    \centering
    \includegraphics[width=1\textwidth]{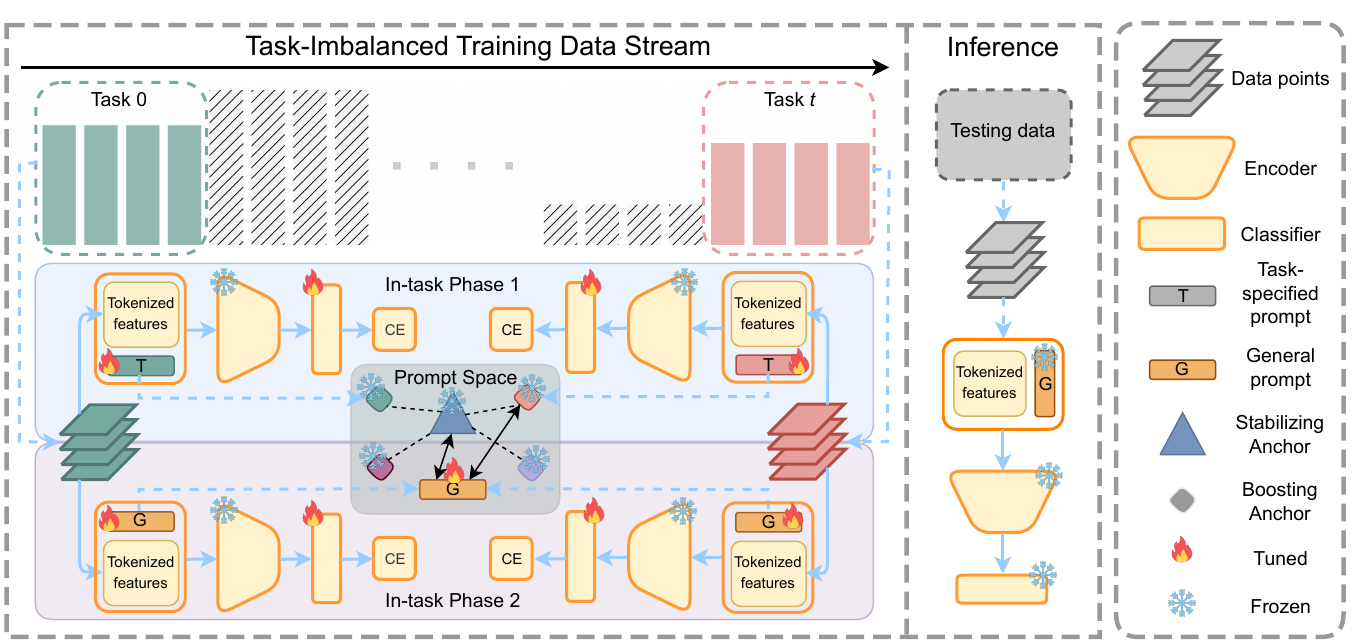}
    \caption{Overview of the proposed dynamic anchored prompting (DAP) for task-imbalanced continual learning. Task-imbalanced training data stream represents the sequential arrival of data from Task 0 to Task \textit{t}. There are two phases for each task. In in-task phase 1, the initialized task-specific prompt learns knowledge related to the current task, serving as a boosting anchor. In in-task phase 2, the general prompt is trained, with the assistance of boosting anchor to ensure plasticity. Meanwhile, the centers of all previously learned task-specific prompts serve as stabilizing anchors, emphasizing stability. It is worth noting that the boosting anchors of past tasks are not stored as the stabilizing anchor is updated in an online manner.}
    \label{fig:method}
    \vspace{-3mm}
\end{figure*}

%% file: tables/table.tex
\begin{table*}[t]
\small
\setlength{\tabcolsep}{4pt}

\centering

\begin{tabular}{c||c c||c c||c c||c c||c c||c c||c}
\hline 
 & \multicolumn{6}{c||}{\textbf{TICL-Cifar100}} & \multicolumn{6}{c||}{\textbf{TICL-ImageNet-R}} \\

\hline 
 & \multicolumn{2}{c||}{Descending} & \multicolumn{2}{c||}{Ascending} & \multicolumn{2}{c||}{Shuffled} & \multicolumn{2}{c||}{Descending} & \multicolumn{2}{c||}{Ascending} & \multicolumn{2}{c||}{Shuffled} & \multicolumn{1}{c}{}\\
\hline


\rule{0pt}{10pt} Method & $A_N$ ($\uparrow$) & $A_L$ ($\uparrow$)  & $A_N$ ($\uparrow$) & $A_L$ ($\uparrow$)& $A_N$ ($\uparrow$) & $A_L$ ($\uparrow$)& $A_N$ ($\uparrow$) & $A_L$ ($\uparrow$)& $A_N$ ($\uparrow$) & $A_L$ ($\uparrow$)& $A_N$ ($\uparrow$) & $A_L$ ($\uparrow$) & Buffer\\

\hline

BiC
& $27.92 $ & $38.02 $
& $36.08 $ & $37.90 $
& $27.11 $ & $33.63 $  
& $21.91 $ & $18.9 $ 
& $13.52 $ & $16.01 $ 
& $16.36 $ & $16.32 $  & $20/\textrm{cls}$\\

PODNET
& $26.48 $ & $23.82 $
& $32.31 $ & $28.24 $
& $28.49 $ & $26.21 $  
& $22.32 $ & $18.90 $ 
& $16.61  $ & $16.20 $ 
& $17.11 $ & $18.70 $ & $20/\textrm{cls}$\\

\hline
EEIL++          
& $31.49 $ & $38.24 $
& $35.93 $ & $37.85 $
& $31.63 $ & $39.31 $  
& $18.50 $ & $17.81 $ 
& $15.76 $ & $15.60 $ 
& $15.79 $ & $16.30 $ & $20/\textrm{cls}$\\
LUCIR++          
& $27.74 $ & $21.26 $
& $42.39 $ & $28.92 $
& $35.62 $ & $25.94 $  
& $18.36 $ & $21.29 $ 
& $8.05 $ & $8.27  $ 
& $15.62$ & $15.62 $ & $20/\textrm{cls}$\\

\hline   
Pre+FT 
& $65.83 $ & $22.02 $
& $19.52 $ & $25.58 $
& $43.30 $ & $33.56 $  
& $40.60 $ & $7.68 $
& $18.22 $ & $21.15 $
& $21.37 $ & $22.62 $
& 0

\\

Pre+iCaRL
& $53.00 $ & $28.73 $
& $41.70 $ & $26.88 $
& $48.62 $ & $31.02 $  
& $48.41 $ & $29.55 $
& $24.40 $ & $29.17 $
& $40.21 $ & $23.02 $ & $20/\textrm{cls}$ \\
\hline

CODA-P
& $\bm{81.91} $ & $58.98 $
& $54.54 $ & $41.84 $
& $60.90 $ & $42.56 $  
& $52.39 $ & $35.21 $
& $28.21 $ & $32.62 $
& $40.02 $ & $34.78 $
& 0
\\

L2-P        
& $66.51 $ & $50.26$
& $53.50 $ & $48.73 $ 
& $51.43 $ & $49.43 $   
& $50.05 $ & $31.72$
& $27.24 $ & $29.42 $
& $30.19 $ & $26.21 $
& 0
\\

Dual-P
& $70.51 $ & $51.79$
& $54.50 $ & $45.72 $ 
& $49.49 $ & $48.82 $   
& $51.47 $ & $31.12 $
& $25.03 $ & $25.42 $
& $34.68 $ & $27.38 $
& 0
\\
\hdashline

\textbf{Ours}   
& $79.09$ & $\bm{61.49} $
& $\bm{56.30}$ & $\bm{55.47 }$
& $\bm{61.43}$ & $\bm{56.12 }$  
& $\bm{58.47}$ & $\bm{40.25}$
& $\bm{31.42}$ & $\bm{36.47}$
& $\bm{43.22}$ & $\bm{36.38}$
& 0
\\ 
\hline
\end{tabular}
\caption{Comparison Results (\%) on TICL-Cifar100 and TICL-ImageNet-R. `Pre' refers to pretraining and `P' stands for prompt. $A_N$ gives the accuracy averaged over tasks, $A_L$ gives the last acccuracy. }
\label{tab:cifar100_main}
\vspace{-15pt}
\end{table*}

%% file: figures/model_eval_pre.tex
\begin{figure}[!t] 
    \centering
    \hspace{-15pt}
    \includegraphics[width=0.9\linewidth]{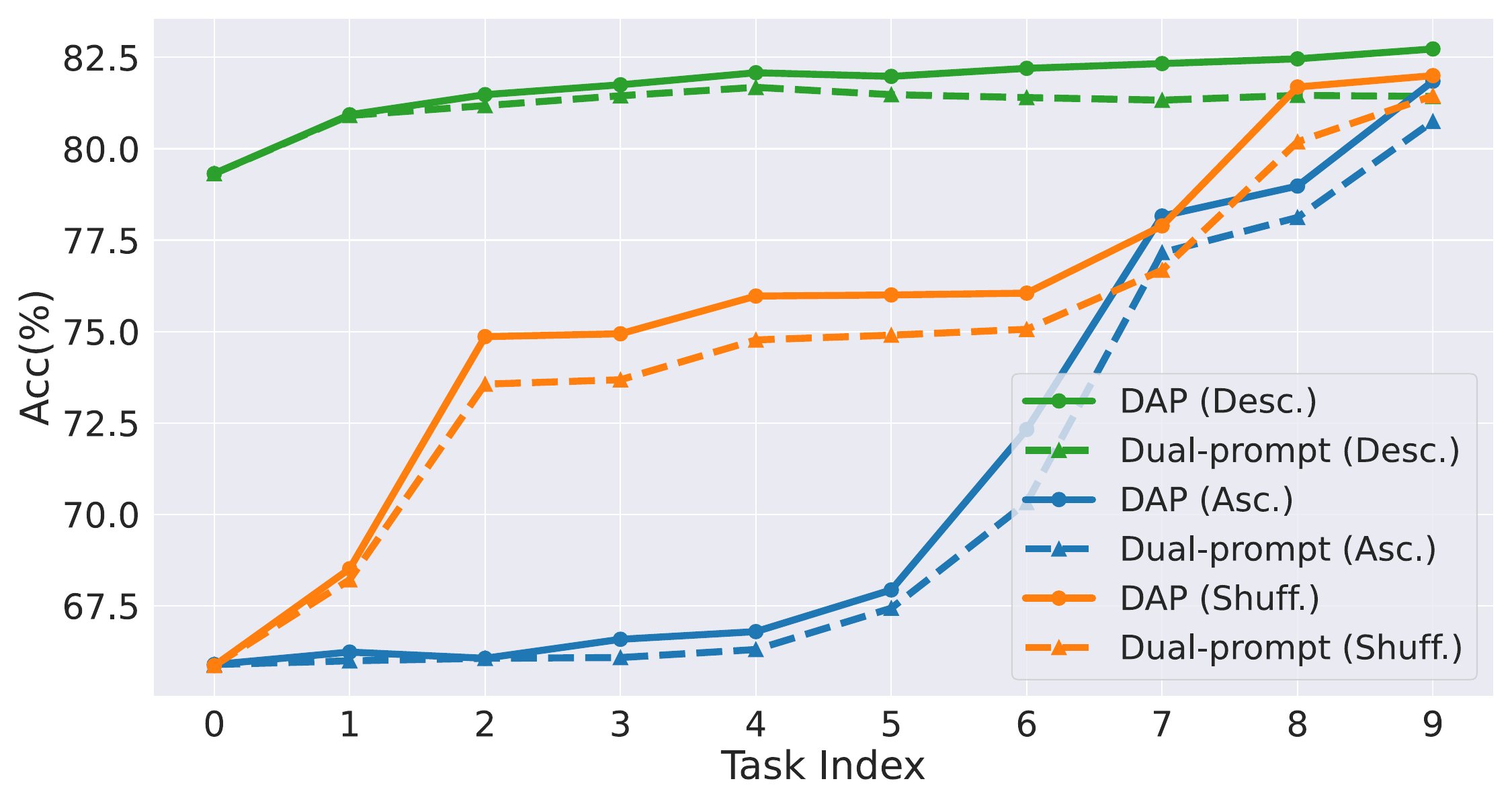}
    \captionsetup{skip=2pt}
    \caption{Performance comparison between DAP and DualPrompt with linear probe.}
    \label{fig:present}
\end{figure}

%% file: tables/abla_g_and_t.tex
\begin{table}[t]
\small
\centering 
\resizebox{0.9\linewidth}{!}
{
\begin{tabular}{l|ccc}
\toprule
 & Desc. & Asc. & Shuff. \\
\midrule
(1) w/ only task-specific prompt  & 64.75 & 52.56 & 52.23 \\
(2) w/ only general prompt  & 66.79 & 50.49 & 47.50 \\
(3) DAP  & \textbf{79.09} & \textbf{56.30} & \textbf{61.43} \\
\bottomrule
\end{tabular}
}
\caption{Ablation on general prompt and task-specific prompt. }
\label{tab:abla_g_t}
\end{table}

%% file: figures/abla_lamda.tex
\begin{figure}[!t]
  \centering
  \hspace{-15pt}
   \includegraphics[width=0.9\linewidth]{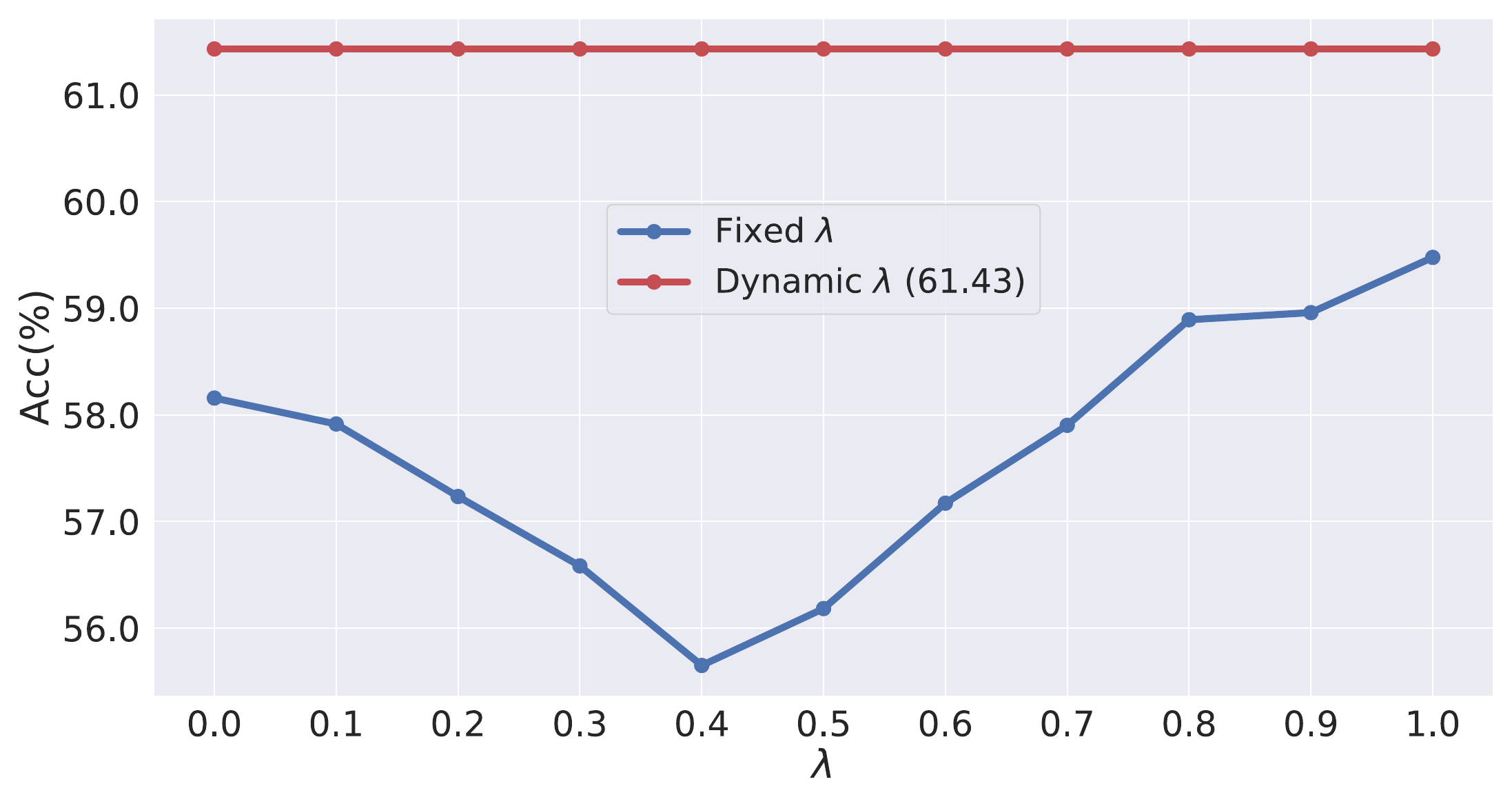}
   \captionsetup{skip=2pt}
   \caption{Ablation on dynamic factor $\lambda$.}
   \label{fig:aba_lamda}

\end{figure}

%% file: tables/abla_anchor.tex
\begin{table}[t]
\small
\centering 
\resizebox{0.83\linewidth}{!}
{
\begin{tabular}{l|ccc}
\toprule

 & $A_N$($\uparrow$) & $F$($\downarrow$) & $P$($\uparrow$) \\

\midrule
         (1) w/ only general prompt   & 47.50 & 18.28 & 63.29 \\
         (2) w/ only boosting anchor     & 58.15 &15.45 & \textbf{70.38} \\
        (3) w/ only stabilizing anchor    & 59.47 &\textbf{~6.19} & 64.86 \\
      (4) DAP   & \textbf{61.43} & ~8.04 & 68.04 \\
\bottomrule
\end{tabular}
}
\caption{Ablation on boosting anchor and stabilizing anchor in shuffled case $A_N$ gives the accuracy averaged over tasks and $F$ gives the average forgetting. $P$ gives the average plasticity.}
\label{tab:abla_anchor}
\end{table}